\documentclass[letterpaper, 10pt, conference]{Docs/ieeeconf}

\usepackage{times}
\usepackage{tabularx}
\usepackage{subcaption}
\usepackage{multirow,multicol, array}
\usepackage[font={small}]{caption}   
\usepackage{graphicx}
\usepackage{dblfloatfix}
\usepackage{wrapfig}
\usepackage{siunitx}
\usepackage{soul}
\usepackage[a-1b]{pdfx}

\let\labelindent\relax
\usepackage{enumitem} 

\usepackage{amsmath,amsthm,amssymb,amsfonts}
\usepackage[titlenumbered,ruled]{algorithm2e}
\usepackage{textcomp}
\usepackage{acronym}
\usepackage{balance}
\usepackage{mdwmath}
\usepackage{amsmath}
\usepackage{bm}
\usepackage{mdwtab}
\usepackage{array}
\usepackage{eqparbox}
\usepackage{cite}

\usepackage{color}
\usepackage{psfrag}
\usepackage{epsfig}
\usepackage{url}

\usepackage{epstopdf}
\usepackage{booktabs}
\usepackage{blindtext}

\usepackage[colorinlistoftodos,prependcaption,textsize=tiny]{todonotes}

\usepackage{lipsum}

\renewcommand{\emph}{\textit}

\newtheorem*{lemma*}{Lemma}

\newtheorem*{problem*}{Problem}

\makeatletter
\newcommand\fs@spaceruled{\def\@fs@cfont{\bfseries}\let\@fs@capt\floatc@ruled
    \def\@fs@pre{\vspace{5\baselineskip}\hrule height.8pt depth0pt \kern2pt}%
    \def\@fs@post{\kern2pt\hrule\relax}%
    \def\@fs@mid{\kern2pt\hrule\kern2pt}%
    \let\@fs@iftopcapt\iftrue}
\makeatother

\IEEEoverridecommandlockouts
\graphicspath{{images/}}

\usepackage[a-1b]{pdfx}

\begin{document}

\title{Development and Testing of a Smart Bin toward \\Automated Rearing of Black Soldier Fly Larvae}

\author{Kevin Urrutia Avila, Merrick Campbell, Kerry Mauck, Marco Gebiola, and Konstantinos Karydis
    \thanks{Kevin Urrutia Avila, Merrick Campbell, and Konstantinos Karydis are with the Dept. of Electrical and Computer Engineering, University of California, Riverside. Email: \{kurru006, mcamp077, karydis\}@ucr.edu. Marco Gebiola and Kerry Mauck are with the Dept. of Entomology, University of California, Riverside. Email: \{marco.gebiola, kerry.mauck\}@ucr.edu}
    \thanks{We gratefully acknowledge the support under a Frank G. and Janice B. Delfino Agricultural Technology Research Initiative Seed Award and a UC MRPI Award. Any opinions, findings, and conclusions or recommendations expressed in this material are those of the authors and do not necessarily reflect the views of the funding agencies.}
}

\maketitle
\thispagestyle{empty}

\begin{abstract}
    The Black Soldier Fly (BSF), \emph{HERMETIA ILLUCENS}, can be an effective alternative to traditional disposal of food and agricultural waste (biowaste) such as landfills because its larvae are able to quickly transform biowaste into ready-to-use biomass. However, several challenges remain to ensure that BSF farming is economically viable at different scales and can be widely implemented. 
    Manual labor is required to ensure optimal conditions to rear the larvae, from aerating the feeding substrate to monitoring abiotic conditions during the growth cycle. This paper introduces a proof-of-concept automated method of rearing BSF larvae to ensure optimal growing conditions while at the same time reducing manual labor. We retrofit existing BSF rearing bins with a ``smart lid," named as such due to the hot-swappable nature of the lid with multiple bins. The system automatically aerates the larvae-diet substrate and provides bio-information of the larvae to users in real time. The proposed solution uses a custom aeration method and an array of sensors to create a soft real time system. Growth of larvae is monitored using thermal imaging and classical computer vision techniques.
    %
    %
    Experimental testing reveals that our automated approach produces BSF larvae on par with manual techniques. 
    
\end{abstract}


\section{Introduction}
Biowaste management is one of the main global challenges of our time, with significant repercussions on human and environmental health related to sanitary issues, pollution of ground water, and emission of greenhouse gases (GHGs). As  global population and consumption rise, biowaste production is also projected to increase significantly~\cite{Kaza2018,Gardner2003}.
Conventional methods of dealing with biowaste, including open dumping in less economically developed countries or landfilling not equipped with means to capture GHGs such as methane, exacerbate the global warming crisis. Besides, landfills release various odors, attract disease vectors, and produce leachates that pollute ground water~\cite{Manfredi2009,Koerner2000}.
Many studies have shown that the insect \emph{HERMETIA ILLUCENS}, commonly known as the Black Soldier Fly (BSF), is able to efficiently turn biowaste into insect biomass that can be used as feed for aquaculture, poultry, livestock and pets~\cite{Muller2017}, and into a compost-like substance known as frass (a mixture of digested biowaste, larval feces and chitin-rich exuviae) that can be used as soil amendment and/or organic fertilizer~\cite{Fonseca2022}. The digestion of biowaste by BSF larvae happens quickly if environmental and substrate conditions are ideal. Therefore, by recycling and upcycling waste, BSF represents an ideal circular economy tool~\cite{SURENDRA202058}.

\begin{figure}[t]
    \vspace{6pt}
    \centering
    \includegraphics[trim={0cm 0cm 0cm 0cm},clip,width=1.0\linewidth,height=6cm]{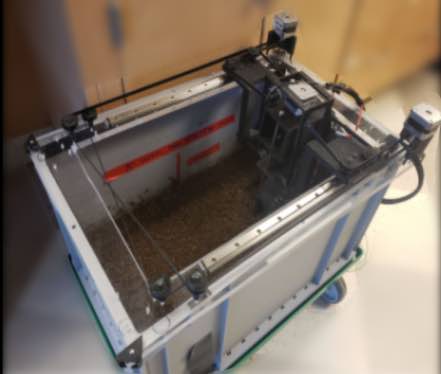}
    \caption{The smart bin design developed in this paper provides an automated solution to both aerating the larval substrate but also monitoring the larval environment. The smart bin enables a hot-swappable approach as it is designed to fit existing commercially-available larval rearing bins.}\label{fig:smartBin} 
    \vspace{-15pt}
\end{figure}

As BSF farming is widely considered a sustainable alternative to traditional biowaste management, there is growing interest to use this form of recycling on a larger scale. Over the past five years, large multinational corporations have been heavily investing in industrial BSF farming and many commercial startups have been established that are also tackling the issues preventing a wider adoption of BSF farming, including automation. This is leading to a large number of patented technological tools that will undoubtedly promote the expansion of BSF farming at the industrial scale. However, there is also a need for open-access and low-cost automated technology to ensure the adoption of BSF farming by the agricultural sector, in particular by small growers and farmers, and by small businesses, in both high- and low-income level countries.
%

Automated solutions need to be able to address basic growth and development needs of the larvae to be considered appropriate. Monitoring of environmental conditions such as temperature~\cite{10.1603/022.038.0347}, relative humidity~\cite{10.1603/EN12054}, substrate moisture~\cite{CHENG2017315}, pH~\cite{10.1371/journal.pone.0202591} and aeration~\cite{ABDUH2022101902} are the most important. Recent attempts at developing an automated solution to BSF rearing have generally been focused on the monitoring of the larval environment aspect. In this light, Internet-of-Things (IoT)~\cite{Wartmann2015} can be leveraged to create a mobile user interface where users can monitor the larval environment in real time~\cite{Van2022} using a built-in display or a web dashboard~\cite{Erbland2021}.
In addition to these aforementioned parameters, it can be important to monitor $NO_{2}$ and $CO_{2}$ too. While these gases are in general not measured, they are relevant as part of studies on life cycle analysis~\cite{SALOMONE2017890}, mass balance~\cite{PARODI2020122488} or global warming potential~\cite{MERTENAT2019173}; hence an automated solution should measure these as well.

\begin{figure}[!t]
\vspace{6pt}
\centering
\includegraphics[trim={0cm 0cm 0cm 0cm},clip,width=0.95\linewidth]{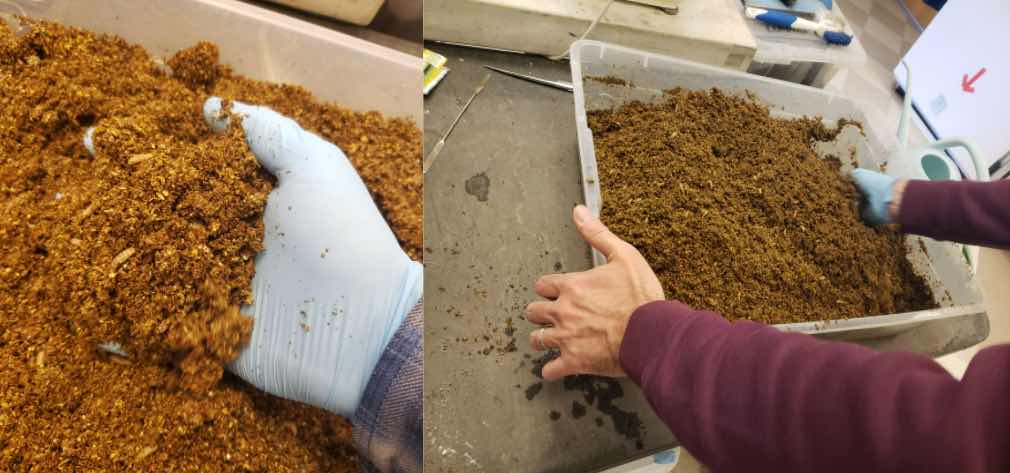}
\caption{Instance of manual larval substrate aeration. The entomologist is required to thoroughly aerate the substrate daily in order to break up the concentration of the larvae. This process can take from $1-5$ min depending on how thoroughly it is being done.}
\label{hand_mix}
\vspace{-18pt}
\end{figure}

However, to the authors' best of knowledge, there has been no development in terms of automation beyond monitoring technology, specifically in terms of both aeration and actuation based on collected bio-data from the larvae. As such, the larval environment still requires an element of aeration as a next step in the automation process. 
The aeration of the substrate requires breaking concentration areas of the larvae. Typically, this is meticulously done by hand once per day, and requires several minutes per bin~\cite{Chapin2013}. 
Introducing an automated solution requires breaking up the concentration of the larvae without any effort from the user. 

This paper seeks to develop an automated solution to BSF rearing (Fig.~\ref{fig:smartBin}) addressing two needs: 1) aerating the larval substrate which is currently a labor-intensive manual process (Fig.~\ref{hand_mix}), and 2) monitoring the environmental conditions of the substrate (temperature, humidity, pH, GHGs) and making sure they are adequate for larvae growth. Our proposed solution, the smart lid (Fig.~\ref{fig:smartBin}), can also monitor the BSF's growth through computer vision techniques and thermal imaging. The smart lid is designed to fit existing commercially-available BSF rearing bins and to be re-usable across bins in a hot-swappable fashion~\cite{Mahmood2021}. 


\section{Technical Approach}\label{sec:approach}

The proposed smart lid for automated BSF larvae rearing hinges on three critical aspects:
1) monitoring the overall growth in the larvae population using computer vision techniques,
2) aerating and mixing the larvae-containing substrate with an automated mechanism, and 
3) logging multiple sensor readings to control the larvae environment.

The proposed solution builds upon existing tools used by entomologists, i.e. a plastic bin used for the rearing of larvae (here we use the bins manufactured by OFERA, Austria), and retrofits it with the designed smart lid to enable automated larvae rearing. The smart lid is designed to fit existing commercially-available bins and affords re-usability on multiple bins by directly swapping the lid between bins. 

The following sections detail the design choices and constraints that inform the critical aspects of the system, ranging from camera selection and placement to the smart lid's interaction with the user.

\begin{figure}[!t]
\vspace{6pt}
\centering
\includegraphics[trim={0cm 1cm 0cm 1cm},clip,width = 0.5\linewidth]{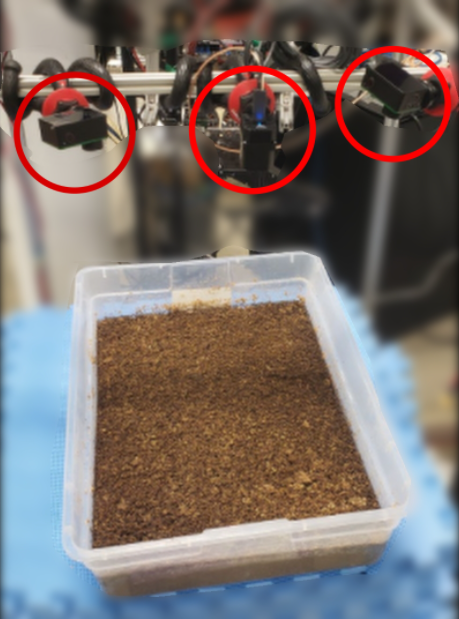}
\caption{Experimental setup to determine appropriate camera type and placement to monitor larvae growth. Cameras from left to right (encircled areas) capture RGB, thermal, and NIR and thermal imaging data, respectively.}
\label{color_setup}
\vspace{-8pt}
\end{figure}

\subsection{Multi-modal Visual Perception for Larvae Growth} \label{perception}

We develop a multi-modal visual perception system to estimate, track, and display the larvae growth to the user. Larvae distribution and growth is estimated via a pixel-mass proxy based on thermal imaging and using classical computer vision techniques, and is presented to the user as drawn contours overlaid on an RGB image.

\begin{figure}[!t]
\vspace{6pt}
\centering
\includegraphics[trim={0cm 0cm 0cm 0cm},clip,width=0.95\linewidth]{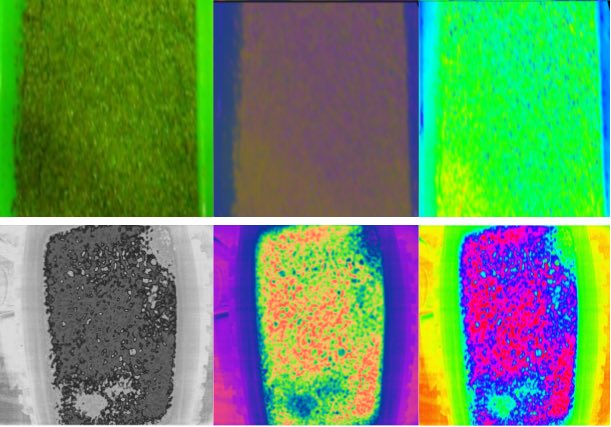}
\caption{NIR (top row) and thermal (bottom row) imaging in BGR (left), YUV (center) and HSV (right) colorspaces. In the thermal images, the HSV magenta regions correspond to clusters of larvae.}
\label{color_spaces}
\vspace{-18pt}
\end{figure}

\begin{figure*}
\vspace{6pt}
\centering
\includegraphics[trim={0cm 0cm 0cm 0cm},clip,width=0.82\paperwidth]{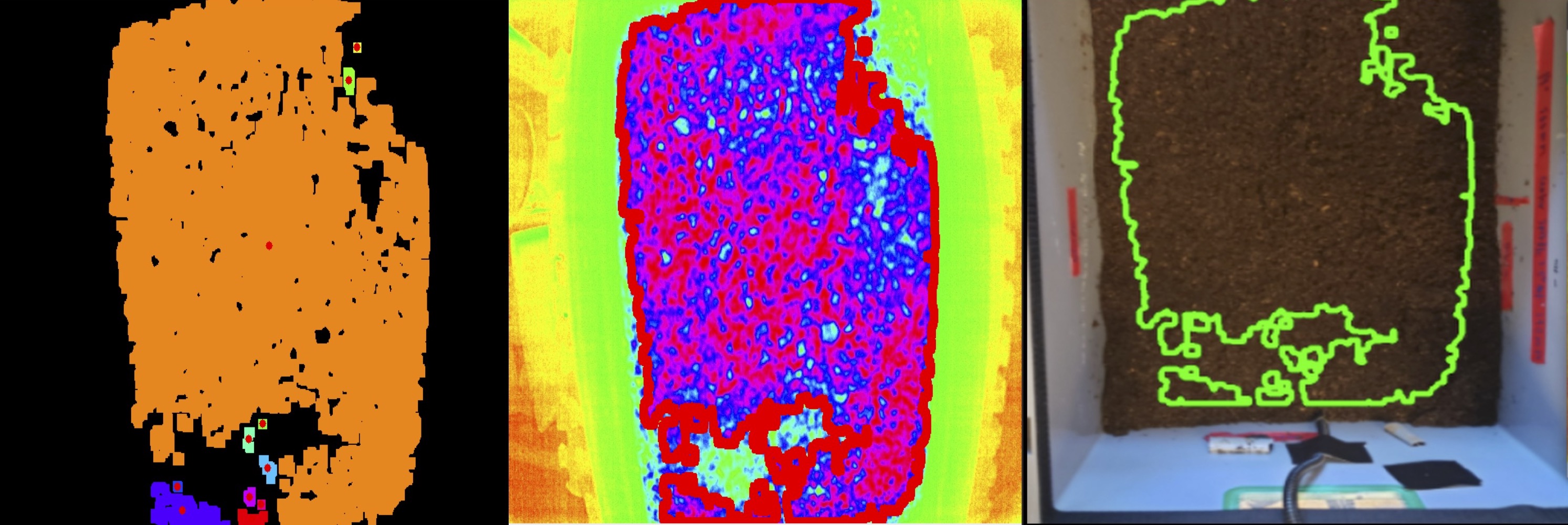}
\caption{Sample visualization provided to the user to estimate larvae growth. (Left) Output of the connected components algorithm displaying a mask of the larval clusters. (Center) HSV thermal image of the larvae (magenta depicts higher concentration) with superimposed outline of the area larvae are within the bin (in red). (Right) superimposed outline of the larval locations within the bin on the RGB image.}
\label{computer_vision_UI}
\vspace{-18pt}
\end{figure*}

In initial phases of development we tested a range of different visual perception sensors to determine appropriate camera type selection and placement. We considered near infrared (NIR), RGB and thermal imaging sensors (Mapir's SURVEY3N-Near Infrared, SURVEY3W- Red+Green+NIR and FLIR's ADK 40640U075-6PAAX, respectively). 
The sensors were fixed at equidistant above a small bin containing larvae and diet (Fig.~\ref{color_setup}), and recorded data throughout larvae gestation, which is approximately $10-15$ days, depending on the type of diet. 
During this experiment the larvae were aerated manually to maintain optimal growth conditions. 


Imaging data were converted into three colorspaces for evaluation: BGR, YUV, and HSV. While not appropriate for tracking and localizing the larvae clusters since they typically occur within the compost and not on its surface, the RGB imaging is useful for displaying user information. NIR imaging was unable to provide useful information about the larvae movement or growth. However, thermal imaging yielded useful insights into the total growth of the larvae population and the locations where the larvae seem to cluster within the bin. Sample NIR and thermal images across the three different color spaces are shown in Fig.~\ref{color_spaces}. We further observed that the HSV colorspace variant of the thermal image can clearly indicate larvae mass concentrations (magenta-colored area). 
Thus, the thermal (in HSV colorspace mode) and RGB cameras were retained for the next iteration of the automated smart lid system and development of the pixel-mass proxy to estimate larvae growth.

Our employed visual perception algorithm for larvae growth estimation begins by first converting thermal imaging into an inferno heat map~\cite{Key2012}. Then, the image is segmented using Otsu's algorithm~\cite{4310076} to separate the larvae from the bin. The segmented heat map is processed with a connected components algorithm~\cite{10.1137/0201010} to determine image statistics such as the centroid and pixel area to be acquired. The mask of the connected components is then used to get the contours of where the larvae are located within the bin. Finally, these contours are superimposed onto both the RGB and thermal images which are then presented to the user (Fig.~\ref{computer_vision_UI}).

Monitoring larval growth in this manner is important because it gives the user the ability to gauge how much the larvae have grown on a day-to-day basis without having to disturb the larval substrate. It also allows to check in a rapid and informative manner how larvae are spatially clustered over the bin, which can in turn trigger when the substrate should be aerated to break up high-concentration larvae clusters.

\begin{figure*}
    \vspace{6pt}
    \centering
    \includegraphics[trim={0cm 0cm 0cm 0cm},clip,width=1.0\linewidth,height=3.95cm]{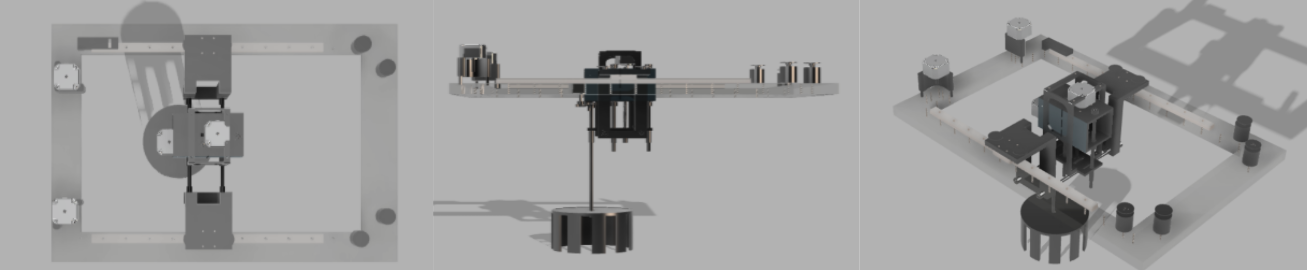}
    \caption{The current gantry design consists of three prismatic joints and a single revolute joint. The purpose of the prismatic joints are to move in the 3D Cartesian space. The revolute joint provides a torque to the spindle at the tool head to aerate the substrate.}\label{fig:mechanism}
    \vspace{-18pt}
\end{figure*}

\subsection{Mechanism Design and Operation for Substrate Aeration}
The proposed aeration mechanism attains the form of a gantry-like system equipped with a specialized spindle as the end-effector for tilling the larvae-impregnated substrate.
(Fig.~\ref{fig:mechanism}). The gantry system contains three chained prismatic actuators to move the spindle in 3D-space. This mechanism's motion is based upon the Core XY structure commonly used in 3D Printer designs~\cite{Yin2018}. This paradigm allows the primary tool carriage to move along the X and Y axes while keeping the motors stationary on the base platform. Placing the motors on the platform instead of having the motors mounted on each consecutive linear motion stage reduces the overall system inertia. Reducing inertia is advantageous to our application since the smart lid design needs to be mounted on a plastic BSF rearing bin where excessive vibration could cause fatigue failure to the structure.

While the Core XY motion mechanism provides several mechanical advantages, the control is somewhat more complex than traditional, independent linear motion stages due to the belt-path coupling. The X and Y motors need to operate in a differential fashion to achieve the desired motion. That is, if $\Delta A$ and $\Delta B$ are the differential linear translations of the belts associated with the two motors, then the equivalent differential translation of the tool carriage along the X and Y axes is given by 
    $\Delta X = \frac{1}{2}(\Delta A + \Delta B)$ 
and 
    $\Delta Y = \frac{1}{2}(\Delta A - \Delta B)$. 

First-principles analysis was conducted to select the motors and other critical components of the device. To retain consistency with the manual aeration process, we considered that the spindle needs to deliver amounts of force similar to a set of human fingers moving through the substrate. This force can be approximated using Stoke's Law as the drag force on a sphere moving through a viscous, that is, 
    $F_s = 6\pi \mu R V$. 
The prototype spindle has a total of 8 fingers.

For this analysis, the dynamic viscosity $\mu$ can be taken to be on the order of that of peanut butter (about $250,000$ cps), which roughly resembles the viscosity of the substrate with larvae. The radius $R$ was taken to be $7.5$\;mm (approximately the size of a human finger and the dimension of each tooth in the tilling spindle). The linear velocity $V$ of the spindle moving through the substrate was taken to be comparable to a human manually aerating the system. 

\begin{figure}[!h]
\vspace{-4pt}
\centering
\includegraphics[trim={0cm 0cm 0cm 0cm},clip,width=0.6\linewidth,height=3cm]{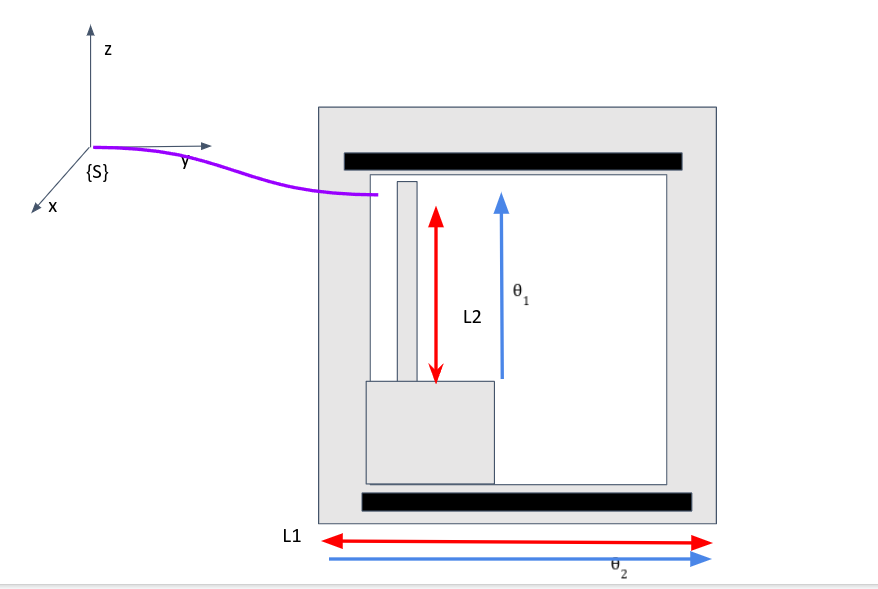}
\includegraphics[trim={0cm 0cm 0cm 0cm},clip,width=0.3\linewidth,height=4cm]{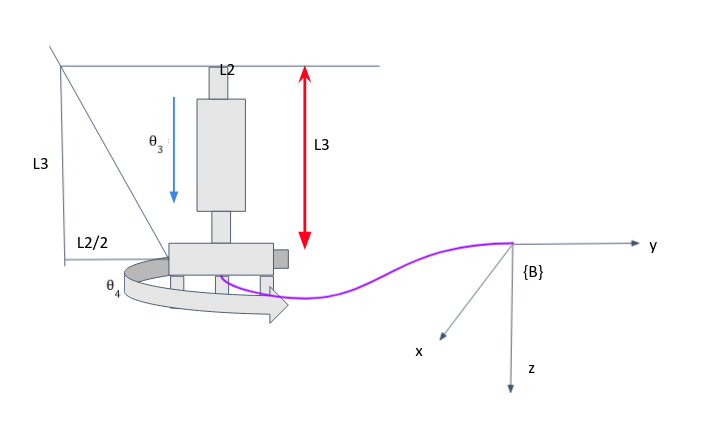}
\includegraphics[trim={0cm 0cm 0cm 0cm},clip,width=1\linewidth,height=4cm]{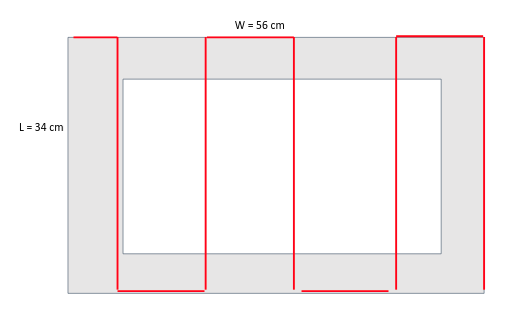}\caption{The mechanism is tasked to aerate the substrate, and therefore it is required to move the spindle through the entirety of the substrate. The raster pattern considered herein can apply. The home position of the mechanism consists of the spindle being to the bottom-leftmost corner of the X-Y plane with the spindle fully retracted out of the substrate. Space and body frames are shown, the difference between them being the Z-axis. The Z-axis of the body frame points down into the substrate.} 
\label{Path}
\vspace{-4pt}
\end{figure}

To determine a lower bound of the desired translational velocity for the spindle, we considered the typical amount of time required in the manual process and the longest non-overlapping path the spindle can follow. 
In detail, the time it takes for an entomologist to thoroughly mix the substrate by hand has been measured at approximately $60$\;sec (for a bin of the size considered herein). For consistency with manual aeration, we desire the mechanism to be able to mix the substrate in the same amount of time as in the manual process. The longest non-overlapping path can be achieved by following a raster motion profile (Fig.~\ref{Path}). In this case, and for the size of the bin considered herein, the spindle needs to travel $1.92$\;m for a single pass. 
Hence, the translational speed of the spindle needs to be at least $0.0032$\;m/s. 

With this analysis in place,\footnote{~We highlight here that this analysis can be directly scaled for different bin sizes and is not specific to the design demonstrated in this paper.}  statics and dynamics computation~\cite{lynch2017modern} can be used to determine the minimum torques required, which in turn is critical for selecting motion generation and transmission components. 
As such, standard GT2 timing belts and pulleys were selected for torque transmission. NEMA-17 Stepper Motors (200 steps/rev, 12V 350mA, 20 N-cm holding torque) were selected for each of the linear motion stages and the spindle. Motor drivers (TB6612) are connected to an Arduino Mega via I$^2$C. Preliminary feasibility testing confirmed that the specific components are appropriate and can operate as desired.


To aerate the substrate, the mechanism begins its path at the home position, as seen in Fig.~\ref{Path}, then drops the carriage and in turn the spindle of the mechanism into the substrate through the use of a lead screw. The spindle begins spinning, and the carriage starts moving according to a desired motion pattern, such as the raster motion specified in Fig.~\ref{Path}.\footnote{~While in this work we consider the raster motion for path planning, other methods can be directly applied as well. These can include predetermined paths (e.g., circular, spiral, meander, feather, etc.), randomly-generated real-time paths, as well as perception-aware ones that are informed by clusters of larvae via thermal imaging (which is enabled by our solution as discussed in Section~\ref{perception}) and perform local spiral paths there.} At the end of the path, the spindle carriage triggers an end stop and returns to the home position and waits for another command to begin again. End stops also connect to the Arduino, which also runs a state machine for operation and logs sensor data.












\subsection{System Integration and Data Collection}


The designed smart lid is effectively a soft real time system that takes input from a real time clock (PCF8523) and decides upon the appropriate action. The system uses a state machine to switch between modes: timer-activated aeration, data collection, and data visualization. In the first mode, the system responds to the system clock and aerates the soil. Typically, the substrate only needs to be aerated once per day. In our experiments, the mechanism has been set for actuation at 11:00 am daily. The mechanism can also be activated on demand, if required.
%
%
%

\begin{figure}[!t]
\vspace{6pt}
\centering
\includegraphics[trim={0cm 0cm 0cm 0cm},clip,width=0.45\linewidth]{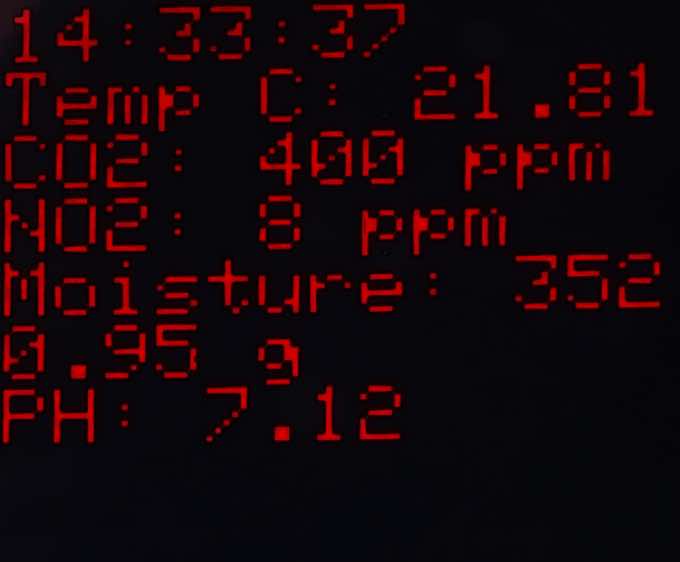}
\includegraphics[trim={0cm 1cm 1cm 0.5cm},clip,width=0.53\linewidth]{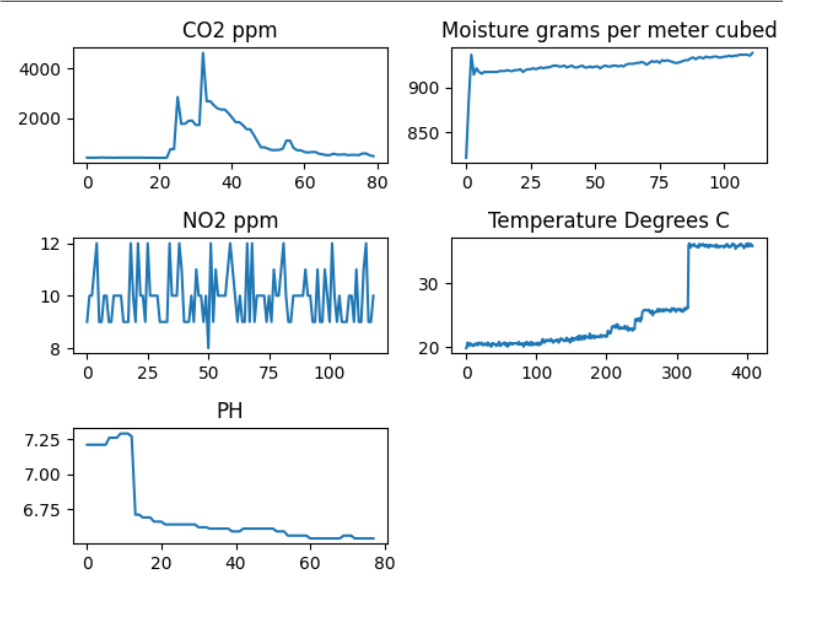}
\caption{The display (left) shows real-time values for the physical states measured by smart lid sensors. (Current values on the display show the larvae environment when the rearing first began.) The system also logs the sensor data over time for later analysis (right).}
\label{lcd_data}
\vspace{-18pt}
\end{figure}

When not activating the aeration mechanism, the system continues the routine process of collecting and logging data to an on-board SD card. The system is also able to store time-stamped data collected from $NO_{2}$, $CO_{2}$, moisture, temperature, and pH sensors. In conjunction with the camera system developed in Section~\ref{perception}, this data can be used to estimate larvae growth. Figure~\ref{lcd_data} shows an example of sensor data post-processing that is available to the user after it is collected by the system.
The data from these sensors can be used to monitor the larvae environment for optimal rearing conditions. As part of future work, we will explore how these sensor inputs can be used to optimize and tune the aeration operations with the spindle and reduce required human intervention and labor.


\section{Experimental Evaluation and Results} \label{system_test}

\begin{figure}[!t]
\vspace{6pt}
\centering
\includegraphics[trim={0cm 0cm 0cm 0cm},clip,width=0.9\linewidth]{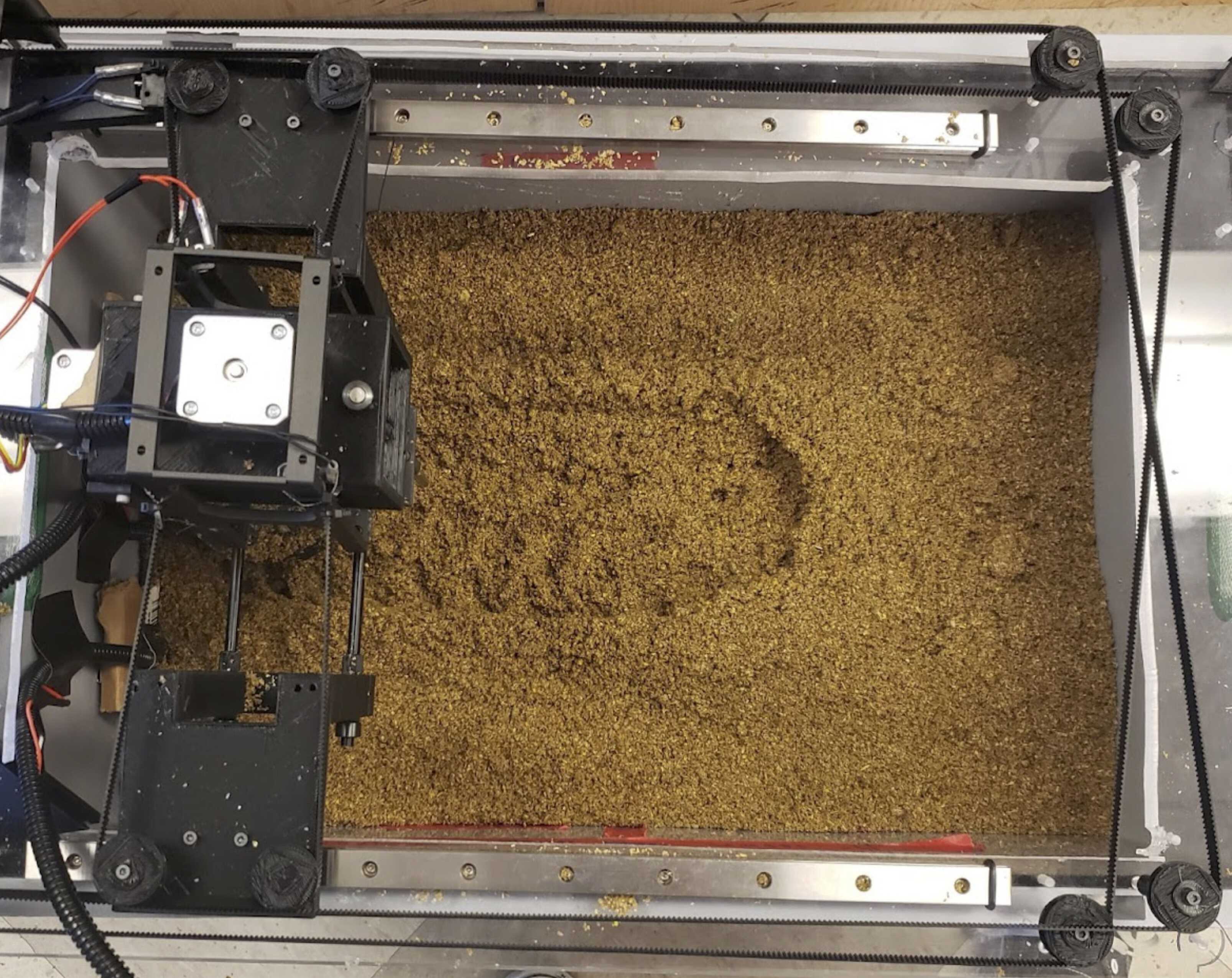}
\caption{Instance of the spindle autonomously aerating the larvae-impregnated substrate at the designated time.}
\label{spindle_op}
\vspace{-9pt}
\end{figure}

After designing and evaluating the individual subsystems for perception and actuation, the integrated automated smart lid was used to rear a batch of larvae. This automated-rearing batch was compared to a manual BSF-rearing system, which was used as the baseline. The substrated for the automated system was aerated by the spindle once per day (Fig.~\ref{spindle_op}). For this analysis, the smart lid was placed upon the bin filled with both the diet and the larvae. Experiments used eight kilograms of Gainesville diet (50\% wheat bran, 30\% alfalfa meal, 20\% corn meal) and 10,000 BSF larvae that are about 5 days old when we start the experiments. 
%
The selected cameras (thermal and RGB) are placed above the bin. Data from the thermal camera were used to compare the larvae growth for the manual and automated BSF rearing trials.



\begin{figure}[!t]
\vspace{6pt}
\centering
\includegraphics[trim={0cm 0cm 0cm 0cm},clip,width=0.99\linewidth]{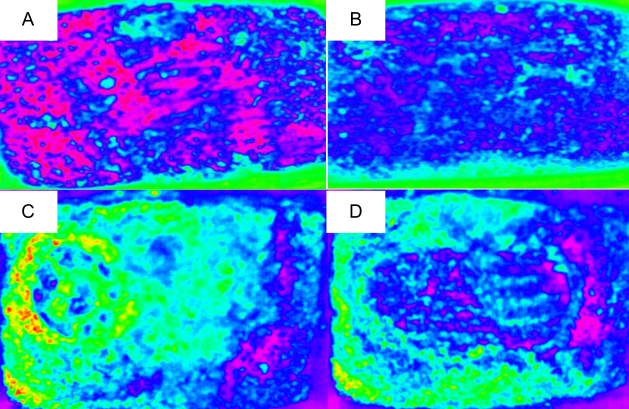}
\caption{In the unaerated soil (A), the larvae clusters are shown in red/magenta. When properly mixed by hand, the clusters should be dispersed (B). Before spindle operation, there are clusters of larvae in red, magenta, and yellow (C). After operation, the clusters are dispersed (D). (Figure best viewed in color.)}
\label{spindle_exp}
\vspace{-18pt}
\end{figure}

Thermal imaging data serve as the means to evaluate the automated aeration mechanism when compared against the manual baseline. Figure \ref{spindle_exp} provides a visualization indicating how aeration affects larvae clusters for both manual and automated processes. 
Before mixing, the thermal images indicate that the larvae are clustered at various areas of the bin; panel (A) corresponds to the manual process while panel (C) corresponds to the automated process. After mixing manually (panel (B)), the larvae are pushed away from the center in a uniform manner and the clustering of the larvae is reduced. In panel (C) of Fig.~\ref{spindle_exp}, we can observe there is a large amount of clustering toward the left edge of the bin. After the automated mixing occurs clusters have broken down as shown in panel (D), indicating that larvae were dispersed and the automated aeration works as intended.
%
%
By segmenting and counting the pixels corresponding to the mixed and unmixed regions, we can tell that the spindle is able to aerate 84\% of the substrate.

After the experiments concluded, the thermal imaging data of both the manual hand-mixed and the automated spindle-mixed bins were analyzed with the algorithm described in Section~\ref{perception} to estimate the mixing efficacy. Since it is not practical to count each of the 10,000 larvae in the bin and follow their individual locations within the bin, this algorithm generated a pixel-mass proxy for the larvae distribution before and after aeration. For the hand-mixed bin, 149,295 pixels corresponded to the well-mixed regions. The spindle-mixed bin contained 101,363 pixels corresponding to the well-mixed regions. This suggests that our automated system mixed 67.9\% of the larvae as the manual process.

\section{Discussion And Outlook}

The Black Soldier Fly (BSF) can be an effective alternative to traditional disposal of biowaste because its larvae can quickly transform it into ready-to-use biomass. However, rearing of BSF larvae can require significant manual labor to perform repetitive tasks such as daily substrate mixing, in addition to monitoring abiotic conditions, which are both required to ensure optimal conditions to rear the larvae. 
The repetitive and labor-intensive process of substrate mixing/aeration can serve as a target for automation. To this end, in this work we developed and tested a smart bin solution to jointly perform substrate aeration, monitor abiotic conditions, and store/present data to the user.

\textbf{Contributions and Key Findings:} 
Our integrated approach merged multi-modal visual perception (namely thermal and visible-spectrum imaging) with actuation and mechanism design, and sensor-based system integration to create a prototype capable of automated aeration of the larval substrate while at the same time offering real time data for the user to manage the larvae environment. 
Thermal imaging data help estimate larvae growth via computer vision tools, and also indicate locations of high larvae concentration that need to be broken down via mixing to promote an appropriate growth environment. First-principles analysis helped determine component selection and informed mechanism design in a way that affords scaling to other bin sizes besides the one considered herein. Results demonstrated the efficacy of our developed automated solution which can aerate the compost comparably to manual mixing.


\textbf{Directions for Future Work:} 
Despite the promising initial capabilities and demonstrated feasibility of our smart bin to successfully curate an environment to rear BSF larvae, the current prototype enables several directions for future automation science and engineering research. 
Additional aeration tests as well as further sensor validation can be done with more types of substrates that differ in viscosity, thus revealing if the designed aeration system is better suited to more or less viscous substrates. 
Further iterations of the software system and development of user interfaces along the lines of IoT would help users see trends in the bio-information of the larvae and adjust more quickly. 
We also seek to introduce additional actuation capabilities to help regulate the environmental conditions during gestation, such as inclusion of sprinklers to increase moisture or fans to reduce gas build up, based on sensor readings or direct user input. 
Finally, redesigning and optimizing the spindle mechanism could improve the automated rearing process and is a promising direction for future work. For instance, we consider integrating a second degree of freedom on the tool head to better aerate the substrate. 


\addtolength{\textheight}{-12cm}

\balance
\bibliographystyle{IEEEtran}
\bibliography{BSF}
\end{document}